\documentclass[conference]{IEEEtran}
\IEEEoverridecommandlockouts
\usepackage{multirow}
\usepackage{tikz}
\usetikzlibrary{calc}
\usepackage{cite}
\usepackage{amsmath,amssymb,amsfonts}
\usepackage{algorithmic}
\usepackage{graphicx}
\usepackage{textcomp}
\usepackage{xcolor}
\usepackage{makecell}
\usepackage{booktabs}
\usepackage{tabularx}
\usepackage{array}
\usepackage{enumitem}
\usepackage{adjustbox}
\def\BibTeX{{\rm B\kern-.05em{\sc i\kern-.025em b}\kern-.08em
    T\kern-.1667em\lower.7ex\hbox{E}\kern-.125emX}}
\usepackage{array}
\usepackage{booktabs}
\usepackage{siunitx}
\usepackage{url}
\begin{document}


\title{The Good, the Better, and the Best: Improving the Discriminability of Face Embeddings through Attribute-aware Learning\\
}

\author{
Ana Dias$^{1,2,3}$, João Ribeiro Pinto$^{4}$, Hugo Proença$^{1,2}$, João C. Neves$^{1,3}$\\[4pt]
$^{1}$University of Beira Interior, Portugal \quad $^{2}$IT: Instituto de Telecomunicações \quad $^{3}$NOVA LINCS \quad $^{4}$Amadeus, Portugal\\
ana.margarida.dias@ubi.pt\\
}
\maketitle

\begin{abstract}

Despite recent advances in face recognition, robust performance remains challenging under large variations in age, pose, and occlusion. A common strategy to address these issues is to guide representation learning with auxiliary supervision from facial attributes, encouraging the visual encoder to focus on identity-relevant regions. However, existing approaches typically rely on heterogeneous and fixed sets of attributes, implicitly assuming equal relevance across attributes. This assumption is suboptimal, as different attributes exhibit varying discriminative power for identity recognition, and some may even introduce harmful biases.
In this paper, we propose an attribute-aware face recognition architecture that supervises the learning of facial embeddings using identity class labels, identity-relevant facial attributes, and non-identity-related attributes. Facial attributes are organized into interpretable groups, making it possible to decompose and analyze their individual contributions in a human-understandable manner. 
Experiments on standard face verification benchmarks demonstrate that joint learning of identity and facial attributes improves the discriminability of face embeddings with two major conclusions: (i) using identity-relevant subsets of facial attributes consistently outperforms supervision with a broader attribute set, and (ii) explicitly forcing embeddings to \textit{unlearn} non-identity-related attributes yields further performance gains compared to leaving such attributes unsupervised. 
Additionally, our method serves as a diagnostic tool for assessing the trustworthiness of face recognition encoders by allowing for the measurement of accuracy gains with suppression of non-identity-relevant attributes, with such gains suggesting shortcut learning from redundant attributes associated with each identity.

\end{abstract}


\newcommand{\grp}[1]{\textsc{#1}}

\section{Introduction}


\begin{figure}[t]
\centering
\begin{tikzpicture}
  \node[inner sep=0] (img) {\includegraphics[width=0.9\columnwidth]{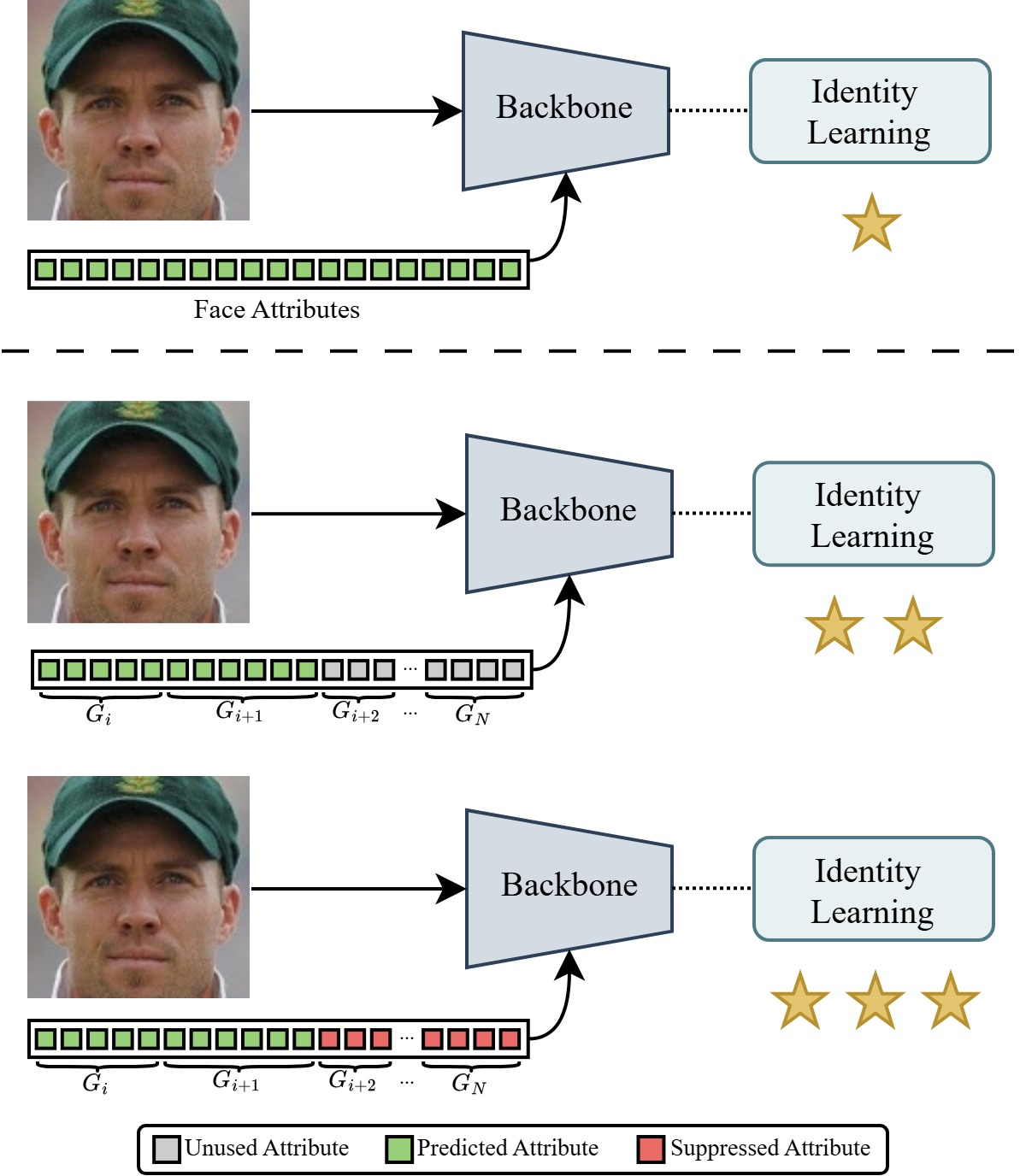}};

\node[rotate=90, anchor=south, font=\fontsize{8.5}{9.5}\selectfont]
  at ($ (img.north west)!{1/6}!(img.south west) + (0mm,3mm) $)
  {(a) Previous Methods};

  \node[rotate=90, anchor=south, font=\fontsize{8.5}{9.5}\selectfont]
    at ($ (img.north west)!{2/3}!(img.south west) + (0mm,3mm) $)
    {(b) Proposed Method};
\end{tikzpicture}
\caption{\textbf{From fixed attribute supervision to selective prediction and suppression.} \textbf{(a)} Prior multi-task face recognition methods use a fixed attribute list as an auxiliary prediction task alongside identity learning. \textbf{(b)} Our approach organizes attributes into groups and controls their influence during training: groups can be predicted or selectively suppressed, while the identity learning objective remains unchanged.}
\label{fig:teaser}
\vspace{-1.5em}
\end{figure}

Despite significant advances over the past decade~\cite{cosface, arcface, magface}, face recognition methods still struggle to correctly identify individuals under factors such as age variation~\cite{ecaf, IEFP}, pose changes~\cite{pose1, pose2}, and occlusion~\cite{occlusion1, occlusion2}. To address these limitations, a common strategy is the use of auxiliary supervision during training, usually with facial attributes, to guide representation learning toward identity-related cues. This is typically achieved within a multi-task learning framework, in which a shared visual encoder is jointly optimized for identity recognition and facial attribute estimation~\cite{Taherkhani, Wang, swinface, faceceptor}. However, existing multi-task face recognition approaches usually treat facial attributes as a single, homogeneous auxiliary objective defined over a fixed attribute set. This formulation neglects two key aspects: (a) facial attributes differ substantially in their relevance and discriminative power for identity recognition, and some may even introduce noise or bias into the learned representation; and (b) actively suppressing certain attributes can be more beneficial than simply leaving them unsupervised. Motivated by these observations, we introduce a face recognition architecture that enhances the discriminability of identity-related features by encouraging the encoder to preserve information associated with identity-relevant facial attributes while ignoring or suppressing information related to attributes that are irrelevant for recognition (Figure~\ref{fig:teaser}). To this end, we adopt a state-of-the-art visual encoder supervised by three complementary loss signals: identity recognition, identity-relevant facial attributes, and identity-irrelevant facial attributes. The first two loss signals are jointly optimized using standard gradient descent to promote the learning of identity-discriminative features. In contrast, the loss associated with identity-irrelevant attributes is combined with a gradient reversal mechanism, encouraging the model to explicitly “unlearn” attribute information that is detrimental to identity recognition. Furthermore, facial attributes are organized into interpretable, region-based groups, enabling a systematic and human-understandable analysis of how different attribute subsets contribute to identity learning. Using the proposed architecture, we demonstrate that jointly learning identity and facial attributes is clearly beneficial to identity-only supervision, but, as hypothesized by us, this increment is suboptimal, since identity-relevant attribute groups yields stronger and more stable improvements than using a heterogeneous mix of attributes. Moreover, we show that explicitly suppressing non-identity-related attributes provides even further performance gains, supporting the hypothesis that actively suppressing weakly relevant attributes can be more effective than simply leaving them unsupervised. While most attribute groups support these conclusions, attributes related to \textit{hair} deviate from the general trend. Despite being non-identity-relevant, its suppression degrades recognition performance, indicating that this region encodes identity-discriminative information. We hypothesize that this is due to the low intra-subject variability of hair-related attributes in common face datasets, and we demonstrate this claim by achieving $\approx 80\%$ verification accuracy solely by using hair-related attributes, without any visual input. These findings expose attribute–identity correlations that are often overlooked in face recognition benchmarks, and demonstrate how our method can be leveraged to identify and analyze such biases. Accordingly, beyond improving recognition performance, the proposed approach provides a useful tool for promoting more transparent and trustworthy face recognition systems by explicitly revealing which attributes contribute to identity discrimination. 

\vspace{0.5em}

In summary, our main contributions are as follows:
\begin{itemize}
    \item We introduce an attribute-aware face recognition architecture that jointly learns identity representations and facial attributes, while explicitly suppressing non-identity-related attribute information via a gradient reversal mechanism.
    \item We experimentally demonstrate that supervising selected identity-relevant attribute groups, together with explicit suppression of non-identity attributes, consistently outperforms joint learning identity with a heterogeneous mix of facial attributes.
    \item We show that selective attribute suppression not only improves recognition performance, but also helps expose shortcut learning in attribute-supervised face recognition, evidencing the potential of our approach for analyzing dataset biases and promoting more transparent and trustworthy recognition systems.
\end{itemize}

\section{State of Art}

\subsection{Multi-task Face Recognition}

Early research on multi-task face recognition explored the use of facial attributes to improve identity learning through auxiliary supervision. Wang et al.~\cite{Wang} proposed a joint optimization framework for face recognition and attribute prediction and reported improved recognition performance, while Taherkhani et al.~\cite{Taherkhani} showed that attributes provide complementary information for identity recognition. Zhong et al.~\cite{Zhong} further showed that face recognition representations capture attribute-related information, consistent with the prior benefits of attribute supervision. More recent frameworks extend this idea to multi-task face perception models with shared backbones and task-specific heads, jointly training face recognition with tasks such as attribute prediction, age estimation, and expression recognition~\cite{faceceptor,swinface}. While these methods demonstrate the effectiveness of joint learning, they typically adopt a fixed attribute list and treat attribute supervision as a single auxiliary objective. Consequently, the effects of different attribute groups, and their selective prediction or suppression, on face recognition performance remain underexplored. Our multitask learning architecture enables controlled prediction and suppression of selected attribute groups and explores their effects on face recognition performance.

\subsection{Adversarial Learning for Invariance}

Gradient Reversal Layers (GRLs)~\cite{GRL} have been adopted in multi-task face recognition to suppress attribute-specific information, most commonly age, from identity representations by reversing the gradient from an auxiliary branch during backpropagation. MTLFace~\cite{ecaf} integrates a recognition discriminator with a GRL~\cite{GRL} to encourage the learned identity feature to be age-invariant, while jointly optimizing age estimation and face recognition objectives. OrdCon~\cite{ordcon} follows a similar strategy and gradually applies a GRL to an age branch to enforce age-invariant identity features. More generally, this same mechanism can be applied to promote invariance to additional facial attributes by attaching an attribute classifier to the identity embedding and training it through a GRL. In our work, we adopt this approach and extend GRL-based suppression beyond age to multiple facial attribute groups.

\begin{figure*}[t] 
    \centering
    \includegraphics[width=1.0\linewidth]{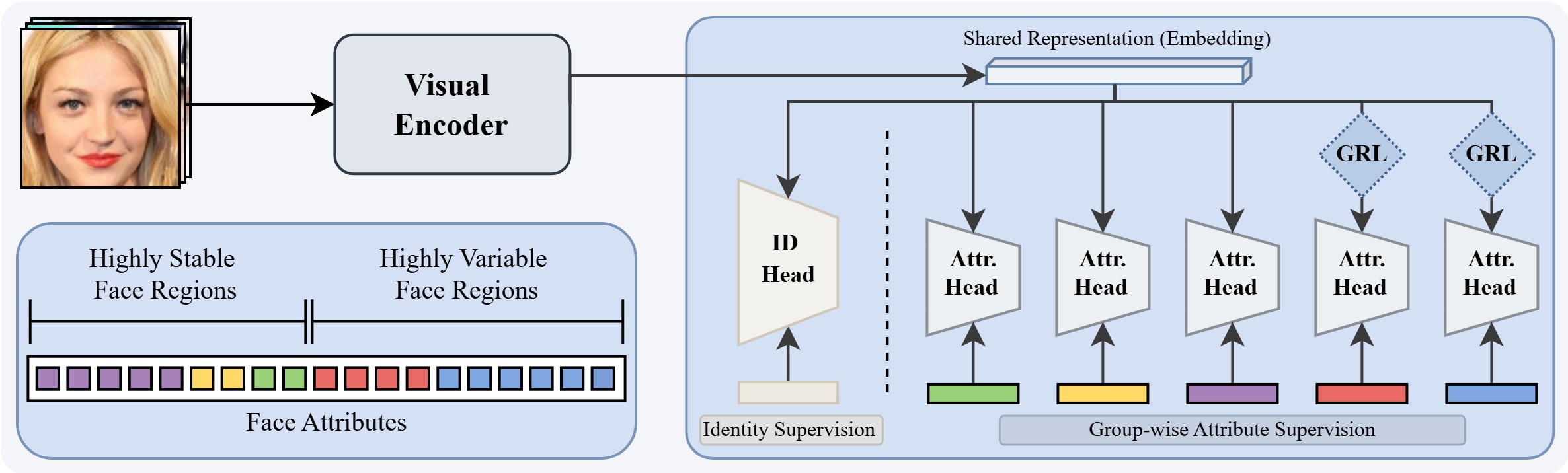}
\vspace{-1.5em}
\caption{\textbf{Overview of the proposed attribute-aware face recognition architecture.} A shared visual encoder produces an identity embedding used for identity recognition and for attribute-based auxiliary supervision. Attribute groups selected for prediction are supervised directly, while groups selected for suppression are connected through gradient reversal layers (GRL) to discourage attribute-specific information in the identity embedding.} 
    \label{fig:methodology}
\vspace{-1.0em}
\end{figure*}

\section{Methodology}
\label{sec:methodology}
This section presents the proposed attribute-aware training architecture for face recognition. We first describe the overall model architecture, followed by the mechanisms used for predicting and suppressing attributes. An overview of the proposed architecture is illustrated in Figure~\ref{fig:methodology}.

\subsection{Network Architecture}
Given an input face image $x$, a shared feature extractor $f_{\theta}(\cdot)$
produces an identity embedding
\begin{equation}
\mathbf{z} = f_{\theta}(x).
\end{equation}

The embedding $\mathbf{z}$ is subsequently processed by an identity classification head $h_{\text{id}}(\cdot)$ and a set of group-specific attribute heads $\{h_g(\cdot)\}$, one for each attribute group $g$. Attribute heads operate in two modes: (i) prediction, where $h_g$ receives $\mathbf{z}$ directly, and (ii) suppression, where $\mathbf{z}$ is first passed through a gradient reversal layer (GRL).


\subsection{Attribute Group Prediction}

We select a set of attribute groups to be trained in prediction mode, denoted as:
\begin{equation}
\mathcal{G}_{\text{pred}} = \{ G_1, \dots, G_K \},
\end{equation}
where each group $G_k$ is a set of attributes. For each group $G_k \in \mathcal{G}_{\text{pred}}$, the corresponding attribute head produces group-specific logits:
\begin{equation}
\hat{\mathbf{a}}_k = h_k(\mathbf{z}),
\end{equation}
where $\hat{\mathbf{a}}_k \in \mathbb{R}^{|G_k|}$ contains one logit per attribute in group $G_k$. The ground-truth binary label of the $i^{th}$  attribute in $G_k$ is denoted by $y_{k,i} \in \{0,1\}$ and the corresponding predicted logit as $\hat{a}_{k,i}$. 

Using the binary cross-entropy (BCE) loss, the attribute prediction loss is defined as:
\begin{equation}
\mathcal{L}_{\text{attr}}
=
\frac{1}{\sum_{k=1}^{K} |G_k|}
\sum_{k=1}^{K}
\sum_{i \in G_k}
\mathrm{BCE}(\hat{a}_{k,i}, y_{k,i}),
\end{equation}
where $|G_k|$ denotes the number of attributes in group $G_k$.

\vspace{0.5em}

\subsection{Attribute Group Suppression}

We select a set of attribute groups for suppression, disjoint from those used
for prediction, denoted as:
\begin{equation}
\mathcal{G}_{\text{sup}} = \{ G_1, \dots, G_M \},
\end{equation}
where each group $G_m$ is a set of attributes. For each group $G_m \in \mathcal{G}_{\text{sup}}$, the corresponding attribute head receives the identity embedding through a gradient reversal layer (GRL):
\begin{equation}
\hat{\mathbf{a}}^{\text{adv}}_m
=
h^{\text{adv}}_m(\mathrm{GRL}(\mathbf{z})),
\end{equation}
where $\hat{\mathbf{a}}^{\text{adv}}_m \in \mathbb{R}^{|G_m|}$ contains one logit
per attribute in group $G_m$. For each attribute indexed by $i \in G_m$, we denote its ground-truth binary label as $y_{m,i} \in \{0,1\}$ and the corresponding predicted logit as $\hat{a}^{\text{adv}}_{m,i}$. 
The adversarial suppression objective is defined as:
\begin{equation}
\mathcal{L}_{\text{adv}}
=
\frac{1}{\sum_{m=1}^{M} |G_m|}
\sum_{m=1}^{M}
\sum_{i \in G_m}
\mathrm{BCE}(\hat{a}^{\text{adv}}_{m,i}, y_{m,i}),
\end{equation}
where $|G_m|$ denotes the number of attributes in group $G_m$.

During backpropagation, the GRL reverses gradients from
$\mathcal{L}_{\text{adv}}$ to $\mathbf{z}$, thereby encouraging the identity
embedding to discard information related to the suppressed attribute groups.

\vspace{0.5em}
\subsection{Multi-task Learning Framework}

We optimize face recognition jointly with the attribute objectives introduced above. The identity classification loss serves as the primary objective, with attribute prediction and suppression losses incorporated as specified below. Identity recognition is supervised using the CosFace loss~\cite{cosface}, defined as:
\begin{equation}
\mathcal{L}_{\text{id}}
= -\frac{1}{N}\sum_{i=1}^{N}\sum_{j=1}^{M}
\log
\frac{
e^{\,r(\cos(\theta_{y_i,i})-m)}
}{
e^{\,r(\cos(\theta_{y_i,i})-m)} + \sum_{j\neq y_i} e^{\,r\cos(\theta_{j,i})}
},
\end{equation}
where $N$ is the batch size, $M$ is the number of identities, $\theta_{j,i}$ denotes the angle between the normalized feature of sample $x_i$ and the normalized prototype of identity $j$, $m$ is the angular margin, and $r$ is the scale factor.

For attribute groups trained in prediction mode, the corresponding heads are optimized using $\mathcal{L}_{\text{attr}}$. For attribute groups trained in suppression mode, we additionally optimize the adversarial loss $\mathcal{L}_{\text{adv}}$.

The overall training objective is given by:
\begin{equation}
\mathcal{L}
=
\mathcal{L}_{\text{id}}
+
\lambda_{\text{pred}} \mathcal{L}_{\text{attr}}
+
\lambda_{\text{adv}} \mathcal{L}_{\text{adv}},
\end{equation}
where $\mathcal{L}_{\text{id}}$ denotes the identity loss, $\mathcal{L}_{\text{attr}}$ the attribute prediction loss, and $\mathcal{L}_{\text{adv}}$ the adversarial loss used for attribute suppression. The weights $\lambda_{\text{pred}}$ and $\lambda_{\text{adv}}$ control the strength of attribute prediction and suppression, respectively.

\section{Experimental Setup\label{sec:expsetup}}


\subsection{Attribute Selection, Grouping, and Experimental Design}

We conduct all experiments using the MAAD-Face dataset~\cite{maad}, which provides large-scale identity annotations together with binary facial attribute labels. Compared to datasets such as CelebA and LFWA~\cite{celebA+LFWA}, MAAD-Face offers greater scale and more reliable identity annotations, making it suitable for analyzing the effect of attribute-based supervision on face recognition performance. 

We organize attribute supervision into region-based groups to study how supervision from different parts of the face contributes to identity recognition, while keeping the analysis focused on interpretable facial areas rather than individual attribute types. Our goal is to contrast sources of supervision that differ in their expected relevance to identity learning. We therefore define five region-based groups, namely \grp{Periocular}, \grp{Mouth}, \grp{Nose}, \grp{Hair}, and \grp{Accessories}. This grouping includes attributes associated with relatively stable facial regions (\grp{Periocular}, \grp{Mouth}, \grp{Nose}) alongside attributes that are more variable or potentially misleading (\grp{Accessories}), as well as attributes that may introduce dataset-specific identity correlations (\grp{Hair}). From the 47 attributes available in MAAD-Face, we retain only those that map directly to these groups, resulting in 19 attributes, as detailed in Table~\ref{tab:attribute-groups}.

\vspace{-1.0em}

\begin{table}[h]
\caption{\textbf{Region-based attribute groups.} Attribute supervision is organized into five region-based groups, namely \grp{Periocular} (P), \grp{Mouth} (M), \grp{Nose} (N), \grp{Hair} (H), and \grp{Accessories} (A). The attributes included in each group are listed.}
\label{tab:attribute-groups}
\centering
\small
\setlength{\tabcolsep}{8pt}
\renewcommand{\arraystretch}{1.12}
\vspace{-0.5em}
\begin{tabularx}{\columnwidth}{@{}l >{\raggedright\arraybackslash}X@{}}
\toprule
\textbf{Groups} & \textbf{Attributes} \\
\midrule \midrule
Periocular (P)  & Bags Under Eyes, Bushy Eyebrows, Arched Eyebrows, No Eyewear, Eyeglasses \\ \hline
\addlinespace[4pt]
Mouth (M)       & Big Lips, Wearing Lipstick \\ \hline
\addlinespace[4pt]
Nose (N)       & Big Nose, Pointy Nose \\ \hline
\addlinespace[4pt]
Hair (H)        & Bald, Wavy Hair, Receding Hairline, Bangs,\\ 
            & Sideburns, Black Hair, Blond Hair, Gray Hair \\ \hline
\addlinespace[4pt]
Accessories (A) & Wearing Hat, Wearing Earrings \\
\bottomrule
\end{tabularx}
\vspace{-0.5em}
\end{table}


Using the framework described in Section~\ref{sec:methodology}, we consider three experimental settings:

\begin{enumerate}[leftmargin=*, itemsep=0.5em]
\item \emph{Single-group prediction.}
We jointly train face recognition and attribute prediction for one group at a time, allowing us to isolate the contribution of each attribute group to recognition performance.

\item \emph{Multiple-group prediction.}
We jointly train face recognition and attribute prediction across multiple groups simultaneously, allowing us to study interactions between different attribute groups and their combined effect on performance.

\item \emph{Multiple-group prediction with suppression.}
We jointly train face recognition with attribute prediction across multiple groups while suppressing selected ones, allowing us to evaluate how reducing information from specific attributes affects recognition performance.
\end{enumerate}


\subsection{Implementation Details}

\noindent \textbf{Training Data.} All models are trained on MAAD-Face~\cite{maad}. Since some attributes contain missing annotations, we restrict training to the largest subset of images for which all selected attributes are defined. This results in approximately 252k training images, which are shared across all experiments. \\ 

\vspace{-2.5mm}
\noindent \textbf{Preprocessing.} For both training and evaluation images, faces are detected and aligned using RetinaFace~\cite{retinaface}, with MTCNN~\cite{mtcnn} used as a fallback. Aligned faces are cropped and resized to a fixed resolution of $112 \times 112$ pixels.\\


\vspace{-2.5mm}
\noindent \textbf{Face Verification Benchmarks.} Face recognition performance is evaluated on four standard verification benchmarks: CALFW~\cite{calfw}, AgeDB~\cite{agedb}, CACD-VS~\cite{cacd}, and ECAF~\cite{ecaf}. All benchmarks follow standard 10-fold cross-validation. \\


\vspace{-2.5mm}
\noindent\textbf{Training Details.} We use a Swin Transformer~\cite{swint}, a state-of-the-art vision encoder tipically used in face recognition~\cite{swinface}, as the shared feature extractor in all experiments. All models are trained using stochastic gradient descent (SGD) with an initial learning rate of $0.01$, reduced by a factor of 10 at epochs 20 and 35. We use a batch size of 512 across all experiments. The CosFace~\cite{cosface} loss is configured with margin $m = 0.35$ and scale $s = 64.0$. We set $\lambda_{\text{pred}}=5$ and $\lambda_{\text{adv}}=2$ based on preliminary validation experiments. Early stopping is applied with a patience of 20 epochs based on validation performance. All experiments are conducted on an NVIDIA A40 GPU.

\section{Results}

\subsection{Effects of Attribute-Based Supervision on Face Recognition}

We evaluate the proposed framework across the three experimental settings defined in Section~\ref{sec:expsetup}, using verification accuracy on the CALFW~\cite{calfw}, AgeDB~\cite{agedb}, CACD-VS~\cite{cacd}, and ECAF~\cite{ecaf} benchmarks, and report the average across them. For ECAF~\cite{ecaf}, we report two protocols: $\langle \text{Adult}, \text{Child} \rangle$ (ECAF$_{a2c}$) and $\langle \text{Child}, \text{Child} \rangle$ (ECAF$_{c2c}$) verification. All configurations are compared against a face recognition baseline trained without attribute supervision, which allows us to isolate the effects of attribute prediction and suppression.\\

\vspace{-0.5em}
\noindent \textbf{Single-group prediction.} Table~\ref{tab:single-groups} reports results from jointly training face recognition and a single attribute group at a time.  This setting isolates each group's individual contribution to recognition performance. Overall, attribute groups associated with relatively stable facial regions, particularly \grp{Periocular} and \grp{Nose}, consistently improve performance over the baseline across most benchmarks, indicating that this supervision shifts the model’s focus toward regions that are reliably informative of identity. \grp{Mouth} provides smaller but generally positive gains, potentially due to higher variability from expression or cosmetics. \grp{Accessories} provides limited and inconsistent improvements, consistent with the expectation that such attributes are weakly related to identity and may introduce noise rather than complementary information. Surprisingly, \grp{Hair} also improves performance on several benchmarks, even though hair-related appearance is more variable and less directly tied to identity than core facial regions. This pattern may reflect dataset-specific correlations between hair attributes and identity, an effect we analyze further in later sections.

\newcommand{\mode}[1]{{\footnotesize #1}}

\begin{table}[h]
\vspace{-0.5em}
\centering
\small
\setlength{\tabcolsep}{2.7pt}
\renewcommand{\arraystretch}{1.2}
\caption{\textbf{Verification accuracy (\%) with single-group attribute supervision.} FR denotes the face recognition baseline trained without attribute supervision. “+P/+M/+N/+H/+A” indicates joint training with prediction of one attribute group (periocular/mouth/nose/hair/accessories). Avg. corresponds to the average verification accuracy across benchmarks.}
\label{tab:single-groups}
\vspace{-0.5em}
\begin{tabular}{@{\hspace{2pt}}l@{\hspace{4pt}}cccccc}
\toprule
Mode & Avg. & ECAF$_{c2c}$ & ECAF$_{a2c}$ & AgeDB & CALFW & CACD-VS \\
\midrule
\mode{FR} & 84.13 & 79.35 & 75.08 & 81.97 & \textbf{86.58} & \underline{97.65} \\
\midrule
\mode{+P} & \underline{84.65} & \underline{79.75} & \underline{76.05} & \textbf{83.57} & 86.22 & \textbf{97.68} \\
\mode{+M} & 84.39 & 79.60 & 75.98 & 83.12 & 85.68 & 97.58 \\
\mode{+N} & \textbf{84.70} & \textbf{80.20} & 75.93 & 83.33 & \underline{86.48} & 97.55 \\
\mode{+H} & 84.52 & 79.55 & \textbf{76.15} & \underline{83.47} & 85.82 & 97.60 \\
\mode{+A} & 84.28 & 78.95 & 75.97 & 83.17 & 85.78 & 97.55 \\
\bottomrule
\end{tabular}
\end{table}


\textbf{Multiple-group prediction and suppression.} Table~\ref{tab:multiple-groups} extends this analysis to multi-group attribute prediction and selective suppression. Combining multiple facial-region groups generally improves performance compared to single-group settings, with the strongest improvements obtained when jointly predicting \grp{Periocular}, \grp{Mouth}, and \grp{Nose}. These regions correspond to different facial areas and provide complementary supervisory information, enabling the model to capture a broader range of facial information that is relevant to identity. However, predicting all attribute groups does not further improve performance. In particular, adding \grp{Hair} and \grp{Accessories} to the \{\grp{Periocular}, \grp{Mouth}, and \grp{Nose}\} configuration degrades average accuracy.

\begin{table}[h]
\vspace{-0.5em}
\centering
\small
\setlength{\tabcolsep}{2.3pt}
\renewcommand{\arraystretch}{1.2}
\caption{\textbf{Verification accuracy (\%) with multiple-group attribute supervision.} FR denotes the face recognition baseline trained without attribute supervision. P/M/N/H/A denote periocular/mouth/nose/hair/accessories groups. “+” denotes prediction of the listed attribute groups, while “--” denotes suppression of the specified group. Avg. corresponds to the average verification accuracy across benchmarks.}

\vspace{-0.5em}
\label{tab:multiple-groups}
\begin{tabular}{@{\hspace{1pt}}l@{\hspace{6pt}}cccccc}
\toprule
Mode & Avg. & ECAF$_{c2c}$ & ECAF$_{a2c}$ & AgeDB & CALFW & CACD-VS \\
\midrule
\mode{FR} & 84.13 & 79.35 & 75.08 & 81.97 & \textbf{86.58} & \textbf{97.65} \\
\midrule
\mode{+PN}         & 84.31 & 79.75 & 75.90 & 82.98 & 85.82 & 97.08 \\
\mode{+PM}         & 84.59 & 79.95 & 76.08 & \textbf{83.40} & 86,08 & 97.42 \\
\mode{+PMN}        & \underline{84.66} & \textbf{80.50} & \underline{76.17} & 83.08 & 86.10 & \underline{97.47} \\
\mode{+PMNHA}      & 84.26 & 79.50 & 75.35 & \underline{83.20} & 85.85 & 97.42 \\
\midrule
\mode{+PMN\,--\,A} & \textbf{84.74} & \underline{80.30} & \textbf{76.53} & 83.12 & 86.08 & \textbf{97.65} \\
\mode{+PMN\,--\,H} & 84.04 & 78.35 & 75.75 & 82.83 & \underline{86.30}  & 96.95 \\
\bottomrule
\end{tabular}
\vspace{-0.5em}
\end{table}

Selective suppression provides further insight: suppressing \grp{Accessories} improves performance, supporting the interpretation that accessory-related supervision is weakly tied to identity and can introduce noise. In contrast, suppressing \grp{Hair} leads to a large performance drop, indicating that the model relies on hair-related information during training, consistent with the single-group setting where \grp{Hair} improved performance despite its higher variability. This reliance likely reflects dataset-specific attribute–identity correlations, which may inflate performance on the evaluation benchmarks and raise concerns about shortcut learning. We also hypothesize that the smaller improvements on CALFW and CACD-VS reflect their more limited age variation, and that benchmarks with larger age gaps, such as AgeDB and ECAF, benefit more from attribute-based supervision.

Overall, these results establish that attribute supervision can improve face recognition, but that its effectiveness depends strongly on how attributes are grouped and selected. Some groups provide complementary identity information, while others encode dataset-specific correlations that the model may learn to rely on. Selective suppression helps reduce this reliance and enables more controlled use of attribute supervision. These findings highlight the need to analyze attribute groups separately and in combination, rather than treating attribute supervision as a single uniform auxiliary signal.

\subsection{Analysis of Identity Predictiveness in Attribute Labels}

The group-wise experiments suggest that attribute supervision can influence recognition not only through complementary identity information, but also through dataset-dependent correlations that the model may exploit. In particular, the strong sensitivity to \grp{Hair} suppression motivates a closer examination of how predictive the attribute labels themselves are of identity. To isolate the signal carried by the attribute annotations, we train classifiers using only attribute labels (no images) under two settings: identification and verification. In both settings, we use the MAAD-Face~\cite{maad} training split and the corresponding attribute annotations.

\vspace{0.5em}

\textit{Identification:} We perform closed-set identity classification from attribute vectors using a linear multiclass SVM and report Rank-1 and Rank-5 accuracy.

\vspace{0.5em}

\textit{Verification:} We construct an attribute-only verification protocol with an equal number of positive/negative pairs, train a linear binary SVM on 1.2M pairs, and evaluate on a disjoint set of 300k pairs.
\vspace{0.5em}

Table~\ref{tab:attrs_only_verif_id} reports identification and verification performance for classifiers trained only on attribute labels, under different single-group and multi-group configurations. Performance is substantially above chance across all groups, reaching 85.31\% verification accuracy when all attribute groups are combined, which shows that the attribute annotations alone contain strong identity-related information. Among individual groups, \grp{Hair} achieves the highest verification accuracy (79.81\%) and the strongest identification results, indicating a particularly strong association between hair-related attributes and identity. These results help interpret the group-wise supervision findings by showing that some attribute groups already carry substantial identity information even without any visual input. In particular, \grp{Hair} is highly predictive in the attribute-only setting, which aligns with its strong impact in the recognition experiments: including \grp{Hair} can boost performance, and suppressing it causes a marked drop. 

Overall, results suggest that attribute supervision effectively improves face recognition performance, while also revealing that some gains may be dataset-dependent and influenced by attribute–identity correlations. The attribute-only analysis confirms that the annotations themselves can be strongly identity-predictive, highlighting the importance of careful attribute selection and control when using attributes for supervision.

\vspace{0.5em}

\begin{table}[t]
\centering
\small
\setlength{\tabcolsep}{2.4pt}
\renewcommand{\arraystretch}{1.12}
\caption{\textbf{Identification and verification results using attribute labels only.} Rank-1/Rank-5 identification and verification accuracy are reported for classifiers trained without images and evaluated across attribute groups and their combinations. Chance performance is shown for reference.}

\label{tab:attrs_only_verif_id}
\vspace{-0.2em}
\begin{tabular}{@{}>{\arraybackslash}p{0.25\columnwidth}ccc@{}}
\toprule
Group & Verif. (\%) & Rank-1 (\%) & Rank-5 (\%) \\
\midrule
Hair        & 79.81 & \underline{1.65} & \underline{6.12} \\
Periocular  & 77.32 & 0.91 & 3.54 \\
Mouth       & 75.26 & 0.53 & 2.20 \\
Nose        & 75.55 & 0.43 & 1.75 \\
Accessories & 75.07 & 0.45 & 1.83 \\
\midrule
\makecell[l]{P,N,M} & \underline{81.47} & 1.56 & 5.68 \\
\addlinespace[2pt]
All Groups  & \textbf{85.31} & \textbf{4.65} & \textbf{14.05} \\
\midrule
\textit{Chance} & \textit{50.00} & \textit{0.02} & \textit{0.10} \\
\bottomrule
\end{tabular}
\vspace{-1.5em}
\end{table}


\vspace{-0.4em}
\section{Conclusion}

In this paper, we introduced an attribute-aware face recognition architecture with controlled attribute supervision during training. Attributes are organized into interpretable region-based groups that can be selectively predicted or suppressed, enabling control over how attribute information influences identity learning. Our experiments show that: 1) selective attribute supervision can consistently improve face recognition performance; 2) combining a small complementary set of attributes is more effective and consistent than supervising all groups jointly; and 3) selective suppression via gradient reversal can also reveal contrasting group contributions, as suppressing accessories-related attributes improves performance, while suppressing hair-related attributes causes substantial drops. Our results also show that annotations themselves carry strong identity-related information even without visual input, highlighting that some gains from attribute supervision may be influenced by dataset-dependent attribute–identity correlations. The proposed approach guides models toward identity-relevant information and can reduce reliance on non-discriminative or dataset-specific signals, improving recognition performance and interpretability.

\vspace{-0.5em}
\section*{Acknowledgments}



This work was funded by the Portuguese Foundation for Science and Technology (FCT) under the PhD grant 2024.02574.BDANA. This work is supported by NOVA LINCS (UID/04516/2025) with the financial support of FCT.IP (DOI: \url{https://doi.org/10.54499/UID/04516/2025}). It was also supported, when eligible, by national funds through FCT under project UID/50008/2025 – Instituto de Telecomunicações (DOI: \url{https://doi.org/10.54499/UID/50008/2025}).

\bibliographystyle{IEEEtran}
\bibliography{references_shorten}    

\end{document}